# Onto Word Segmentation of the Complete Tang Poems

Chao-Lin Liu
Department of Computer Science
National Chengchi University
chaolin@nccu.edu.tw

**Introduction**

The research community agrees on the influences and importance of the Tang[1] poems for the studies of Chinese literature and linguistics,[2] and we have seen discussions about the Tang poems in the community of digital humanities.[3] Chinese is a language that does not place clear markers, e.g., spaces in alphabetic languages, between words in the texts. To study the Chinese contents with digital tools, words are more meaningful units than characters, although using characters is also possible. Unfortunately, it remains difficult to algorithmically segment words in classical Chinese.[4]

Chinese word segmentation (**CWS**) is a key step for computational processing of Chinese texts. CWS for modern vernacular Chinese has achieved good results in international competitions and in practical applications.[5] It is relatively easier to obtain data to train software for CWS of modern Chinese, but there are no known good sources of labelled data for CWS of classical Chinese yet. In fact, even segmenting classical Chinese texts into sentences is an ongoing research topic.[6]

We report results of our attempt to segment words in the Complete Tang Poems (**CTP**). CTP is a representative and arguably the most famous collection of Tang Poems.[7] To study CWS for CTP, we must acquire poems that are segmented by domain experts. At this moment, we use only regulated pentametric octaves (**RPO**) and regulated heptametric octaves (**RHO**) of seven prominent poets.[8] The segmentation task was achieved by five colleagues who have university-level domain knowledge in Chinese poetry.[9] We have to ignore two poems because they include very rare characters that cannot be handled in our programming environment. We also dropped nine poems when our annotators believed that the poems could be segmented in multiple ways or when our annotators were not sure how to segment them. At this moment, we have 2433 segmented regulated octave (**RO**) poems.[10]

---

[1] Tang is a Chinese dynasty that existed between 618CE and 907CE.

[2] See, for instance, (1) Fengju Lo (2005) Design and applications of systems for word segmentation and sense classification for Chinese poems, *Proceedings of the Fourth Conference of Digital Archive Task Force*. (in Chinese); (2) Jeffrey R. Tharsen (2015) *Chinese Euphonics: Phonetic Patterns, Phonorhetoric and Literary Artistry in Early Chinese Narrative Texts*, PhD Dissertation, University of Chicago, USA; (3) Thomas J. Mazanec (2017) *The Invention of Chinese Buddhist Poetry: Poet-Monks in Late Medieval China*, PhD Dissertation, Princeton University, USA; and (4) John Lee, Yin Hei Kong, and Mengqi Luo (2018) Syntactic patterns in classical Chinese poems: A quantitative study, *Digital Scholarship in the Humanities*, 33(1):82–95.

[3] See, for instance, (1) the fourth example in footnote 2 and (2) Chao-Lin Liu, Shuhua Zhang, Yuanli Geng, Huei-ling Lai, and Hongsu Wang (2017) Character distributions of classical Chinese literary texts: Zipf's law, genres, and epochs, *Proceedings of the 2017 International Conference on Digital Humanities*, 507–511.

[4] Shilei Huang and Jiangqin Wu (2018) A pragmatic approach for classical Chinese word segmentation, *Proceedings of Eleventh International Conference on Language Resources and Evaluation*, 1161–1168.

[5] Hao Zhou, Zhenting Yu, Yue Zhang, Shujian Huang, Xi–u Dai, and Jiajun Chen (2017) Word-context character embeddings for Chinese word segmentation, *Proceedings of the 2017 Conference on Empirical Methods in Natural Language Processing*, 760–766.

[6] Chao-Lin Liu and Yi Chang (2018) Classical Chinese sentence segmentation for tomb biographies of Tang dynasty, *Proceedings of the 2018 International Conference on Digital Humanities*, 231–235.

[7] Dingqiu Peng (1645-1719) et al. eds. *Quan Tang Shi* 全唐詩 (The Complete Tang Poems). Beijing: Zhonghua Book Company (2003). (in Chinese)

[8] notes: (1) Regulated pentametric octaves and regulated heptametric octaves mean 五言律詩 and 七言律詩, respectively. (2) An octave poem has eight lines. (3) The poets are Yuan Zhen(元稹), Li Shangyin(李商隱), Li Bai(李

| pattern | frequency | percentage | pattern | frequency | percentage |
|---|---|---|---|---|---|
| 2-2-2-1 | 3009 | 15.5% | 2-2-1 | 4861 | 25.0% |
| 2-2-1-2 | 4568 | 23.5% | 2-1-2 | 6317 | 32.5% |
| 2-2-3 | 214 | 1.1% | 2-3 | 94 | 0.5% |

Table 1. Frequencies and percentages of the most frequent patterns in 19464 lines

There is a popular belief, among experienced researchers and readers of classical Chinese poems, about the word boundaries in the RPO and RHO poems.[11] A sentence in an RPO poem contains five characters. They can be segmented into two patterns: 2-2-1 or 2-1-2, where 2-2-1 indicates that a sentence is segmented into a two-character word, a two-character word, and a one-character word. Similarly, a sentence in an RHO poem can be segmented into two patterns: 2-2-2-1 and 2-2-1-2.

We have 19464 lines in the 2433 RO poems, and found that 96.5% of the lines followed the aforementioned expectation.[12] Table 1 shows the details. The most common exceptions are due to place or person names in poems, and, in such cases, we observed 2-3 or 2-2-3 patterns. They represent 1.6% of the lines.

## Weighted Pointwise Mutual Information (**PMI**)

The simplest version of a CWS problem is to determine whether we should segment a sequence of three characters, say XYZ, into XY-Z or X-YZ.[13] We compute the PMI for the competing candidates, and choose the bigram that has a larger PMI. Namely, we choose XY-Z if PMI(XY) is larger than PMI(YZ). The PMI of a bigram, *AB*, is defined below.[14]

$$\text{PMI}(AB) = \log_2 \left( \frac{\Pr(AB)}{\Pr(A)\Pr(B)} \right)$$

Normally, we use labelled data in machine learning research for training. However, we have only 2433 segmented poems, and will use them to evaluate our methods for segmentation. Hence,

---

白), Du Mu(杜牧), Du Fu(杜甫), Bai Jyuyi(白居易), Wei Yingwu(韋應物).
[9] Information about our annotators remains concealed for anonymous submission to DH 2019.
[10] 1427 regulated pentametric octaves and 1006 regulated heptametric octaves
[11] Fengju Lo (2005) Design and applications of systems for word segmentation and sense classification for Chinese poems, *Proceedings of the Fourth Conference of Digital Archive Task Force*. (in Chinese)
[12] An octave poem has eight lines, so we have 2433×8=19464 lines.
[13] **Here, an English letter represents a Chinese character.** XYZ could represent "依山盡" in "白日依山盡", which is a line in one of Li Bai's poems. Our question is whether we should segment XYZ into XY-Z or X-YZ, i.e., should we choose 依山-盡 or 依-山盡.
[14] Christopher D. Manning and Hinrich Schütze. *Foundations of Statistical Natural Language Processing*. The MIT Press. 1999.

we cannot use 2433 poems for training, and have to train PMI values with other poems to segment the lines based on domain knowledge or heuristics.

Consider a line, JKLMNOP, in an RHO poem.[15] Although we are not sure of the correct segmentation, we can assume that this line may follow either the 2-2-1-2 or 2-2-2-1 pattern as we explained above, and record the occurrences of JK, LM, NO, and P (if 2-2-2-1) or the occurrences of JK, LM, N, and OP (if 2-2-1-2). Since we are not really sure of the correct pattern for a line, we can only assign different weights to JK, LM, N, NO, OP, and P based on certain assumptions. We can then use the weights for unigrams and bigrams to estimate the probability values.

In this running example, we are more confident that we will see the occurrences of JK and LM than the occurrences of NO and OP, so it is reasonable to assign larger weights to JK and LM than to NO and OP. Under the current assumption, KL is unlikely to form a bigram, but we may choose to assign a small weight to its occurrences. This can be done in many different ways, and we will report technical details in DH2019.

We can compute the PMI for bigrams with 38580 CTP poems that contain only five or seven characters in their lines. Since there are less than 7500 distinct characters in CTP, we hope that having more than 2.22 million characters in 38580 poems will provide statistics about the PMI values with reasonable reliability.

We do not use the poems of the poets whose poems are in our segmented poems for training PMI values because we use the segmented poems as the test data. Hence, it is possible to encounter unseen unigrams and bigrams at test time. In these cases, we adopt a basic smoothing procedure to estimate the unseen instances.

**Segmentation with Weighted PMI**

We can measure the quality of segmentation decisions in different ways. The most common measurement is the precision rate, recall rate, and $F_1$ measure.[14] Denote the segmentation decisions for a line as an array of either N (for not segmenting) or P (for segmenting). Consider a seven-character line. There are six positions to place segmentation markers between the characters, and the correct decision for a line of 2-2-2-1 pattern is NPNPNP. An array of NPPNPP will produce a 2-1-2-1-1 pattern, and resulting in 2/4, 2/3, and 4/7 in precision, recall, and $F_1$, respectively.

Among our 2433 segmented poems, we can find 2009 poems that contain only lines that conform to the four patterns.[16] For these patterns, the segmentation problem is boiled down to choosing either 2-1-2 or 2-2-1 for RPO poems and either 2-2-1-2 or 2-2-2-1 for RHO poems. Hence, we expect to achieve more favorable results when we use these 2009 poems as the test data.

Using this prior information in our segmenter, we produce only NPPN or NPNP decisions for RPO poems and NPNPPN or NPNPNP for RHO poems. Running our segmenter with the 2433

---

[15] A RHO poem contains seven characters, and, again, we use an English character to represent a Chinese character.
[16] These four patterns are 2-2-1 and 2-1-2 for regulated pentametric octaves and 2-2-2-1 and 2-2-1-2 for regulated heptametric octaves.

| groups of test data | 2433 poems | | | 2009 poems | | |
|---|---|---|---|---|---|---|
| measurements of quality | $F_1$ | WR | PSP | $F_1$ | WR | PSP |
| completely random | 49.7% | 19.8% | 0.00% | 50.1% | 20.1% | 0.00% |
| 4 patterns, but random | 78.3% | 69.3% | 0.12% | 78.7% | 70.0% | 0.40% |
| 4 patterns + PMI | 89.6% | 85.2% | 11.7% | 90.3% | 86.3% | 14.2% |
| 4 patterns + PMI + parallelism | 89.9% | 85.7% | 17.8% | 90.7% | 86.9% | 21.5% |

Table 2. Quality of Segmentation for different combinations of strategies and datasets (WR: percentage of word recovery; PSP: percentage of perfectly segmented poems)

poems, we achieved **89.6%** in $F_1$. When experimenting on the 2009 poems, we achieved **90.3%** in $F_1$.[17]

### More Demanding Measurements

An $F_1$ of about 90% is an encouraging accomplishment, but it is somewhat clement. A correct segmentation decision does not guarantee the identification of a word. We need two correct decisions on both sides of a word to find the word. Hence, a more practical measure is to calculate the proportions of words recovered by our decisions. Running our segmenter with the 2433 poems and the 2009 poems, we recovered **85.2%** and **86.3%** of the words, respectively.[17]

Is 90% in $F_1$ easy to achieve? Is recovering 86% of the words effortless? If we randomly choose one from NPPN and NPNP for RPO poems and one from NPNPPN and NPNPNP for RHO poems, we achieved only **78.3%** in $F_1$ and recovered only **69.3%** of the words for the 2433 poems.[18]

### Contributions of More Domain-Dependent Information

Domain-dependent information is instrumental. If we do not employ the presumption about the four patterns and make a random guess for every segmentation decision, we would see **49.7%** in $F_1$ and recovered **19.8%** of the words for the 2433 poems.[19]

Researchers found that it is more likely for the middle four lines in RO poems to parallel.[20] Taking this factor into consideration, our segmenter can segment a pair of lines in the same pattern. We achieved **89.9%** in $F_1$ and recovered **85.7%** words for the 2433 poems.[21]

The resulting improvements in $F_1$ and word recovery are not very impressive by adding parallelism into consideration. We have to consider an even stricter measurement: the percentage of perfectly segmented poems. Considering the number of correct decisions needed to perfectly segment a poem, this measurement is very challenging. After including the parallelism factor, we raised this percentage from **14.2%** to **21.5%** for the 2009 poems (see the last two rows of Table 2).

---

[17] Statistics for this method of segmentation is shown in row "**4 patterns + PMI**" in Table 2.
[18] Statistics for this method of segmentation is shown in row "**4 patterns, but random**" in Table 2.
[19] Statistics for this method of segmentation is shown in row "**completely random**" in Table 2.
[20] John Lee, Yin Hei Kong, and Mengqi Luo (2018) Syntactic patterns in classical Chinese poems: A quantitative study, *Digital Scholarship in the Humanities*, 33(1):82–95.
[21] Statistics for this method of segmentation is shown in row "**4 patterns + PMI + parallelism**" in Table 2.

## Concluding Remarks and Recent Progress with Word Vectors

Table 2 summarizes the experimental results that we have discussed and observed in experiments that considered different combinations of segmentation methods and types of test data. Using the weighted PMI scores and adopting appropriate domain knowledge help the segmenter achieve better results. Our results are based on 2433 poems of seven famous poets. It is intriguing to replace the PMI scores with the cosine similarity that we can compute with the word vectors,[22] but we only observed some poor results in few preliminary explorations. More recently, we have increased the amount of labelled data significantly, and were able to apply deep learning, including the LSTM units (long short-term memory) in our classifiers for the CWS task. With the increased data, we have boosted the performance noticeably, and we shall discuss these latest results at DH 2019.

## Acknowledgments

The research was supported in part by contracts MOST-104-2200-E-004-005-MY3 and MOST-107-2221-E-004-009-MY3 of the Ministry of Science and Technology of Taiwan and in part by projects 107H121-06, 107H121-08, 108-H121-06, and 108H121-08 of the National Chengchi University. The segmentation task was carried out by five assistants who major in Chinese Literature. We thank Yi-lin Dai, Nai-An Fu, Wei-Ting Huang, and Shuo-Feng Tsai of the University of Taipei and Yu-Ching Song of the National Taipei University.

**The main body of this proposal includes 1491 words, excluding the title, the tables, the footnotes, and this statement.**

---

[22] Tomas Mikolov, Ilya Sutskever, Kai Chen, Greg S. Corrado, and Jeff Dean (2013) Distributed representations of words and phrases and their compositionality, *Advances in Neural Information Processing Systems* 26, 3111–3119.